\documentclass{article}
\usepackage{ijcai22}

\usepackage{times}

\usepackage{soul}
\usepackage{url}
\usepackage[hidelinks]{hyperref}
\usepackage[utf8]{inputenc}
\usepackage[small]{caption}
\usepackage{graphicx}
\usepackage{amsmath}
\usepackage{booktabs}
\usepackage{amssymb}
\usepackage{graphicx}

\title{Unsupervised Multi-Modal Medical Image Registration via Discriminator-Free Image-to-Image Translation}

\author{
Zekang Chen$^1$\and
Jia Wei$^1$\footnote{Corresponding Author}\And
Rui Li$^2$\\
\affiliations
$^1$School of Computer Science and Engineering, South China University of Technology, Guangzhou, China\\
$^2$Golisano College of Computing and Information Sciences, Rochester Institute of Technology, Rochester, NY 14623, USA\\
\emails
csjwei@scut.edu.cn
}

\begin{document}

\maketitle

\begin{abstract}
In clinical practice, well-aligned multi-modal images, such as Magnetic Resonance (MR) and Computed Tomography (CT), together can provide complementary information for image-guided therapies. Multi-modal image registration is essential for the accurate alignment of these multi-modal images. However, it remains a very challenging task due to complicated and unknown spatial correspondence between different modalities. In this paper, we propose a novel translation-based unsupervised deformable image registration approach to convert the multi-modal registration problem to a mono-modal one. Specifically, our approach incorporates a discriminator-free translation network to facilitate the training of the registration network and a patchwise contrastive loss to encourage the translation network to preserve object shapes. Furthermore, we propose to replace an adversarial loss, that is widely used in previous multi-modal image registration methods, with a pixel loss in order to integrate the output of translation into the target modality. This leads to an unsupervised method requiring no ground-truth deformation or pairs of aligned images for training. We evaluate four variants of our approach on the public Learn2Reg 2021 datasets~\cite{hering2021learn2reg}. The experimental results demonstrate that the proposed architecture achieves state-of-the-art performance. Our code is available at https://github.com/heyblackC/DFMIR.
\end{abstract}

\section{Introduction}
Different medical image modalities, such as Magnetic Resonance Imaging (MRI) and Computed Tomography (CT), show unique tissue features due to the acquisition with different scanners and protocols. Multiple image modalities can be fused to provide combined information, which is known as the process of multi-modal image fusion~\cite{pielawski2020comir}. A wide variety of clinical applications rely on the fusion of multi-modal data, e.g., preoperative planning and image-guided radiotherapy~\cite{oh2017deformable}. Since the images need to be aligned first through registration for image fusion, it is of great importance to establish anatomical correspondences among images of different modalities by using multi-modal image registration.

Learning-based registration seeks to predict deformation fields directly from a pair of images by maximizing a predefined similarity metric~\cite{fan2019adversarial}. Supervised or semi-supervised learning strategies use ground-truth deformation fields or segmentation masks in the training phase, and may suffer from the lack of data labeling~\cite{uzunova2017training,hu2018weakly}. Since it is extremely time-consuming and laborious to label registration data even for specialists, unsupervised methods have been proposed to overcome this limitation solely by maximizing the image similarity between the target image and the source image. However, the performance of unsupervised methods is highly dependent on the choice of cross-modal similarity metrics. Generally, widespread similarity metrics like the sum of squared differences (SSD) and normalized cross correlation (NCC), which are well-suited for mono-modal registration problems~\cite{balakrishnan2019voxelmorph,de2017end}, perform badly in a multi-modal setting. Typically, unsupervised multi-modal registration approaches use Normalized Mutual Information (NMI) and Modality-Independent Neighbourhood Descriptor (MIND)~\cite{maes2003medical,heinrich2012mind}. Since NMI, as a global metric, only measures statistical dependence between two entire images, it is difficult to use it for local image alignment. MIND, on the other hand, is a patch-based image similarity metric, which tends to suffer from severe image deformations and cannot achieve global alignment.

\begin{figure}
	\centering
	\includegraphics[scale=0.47]{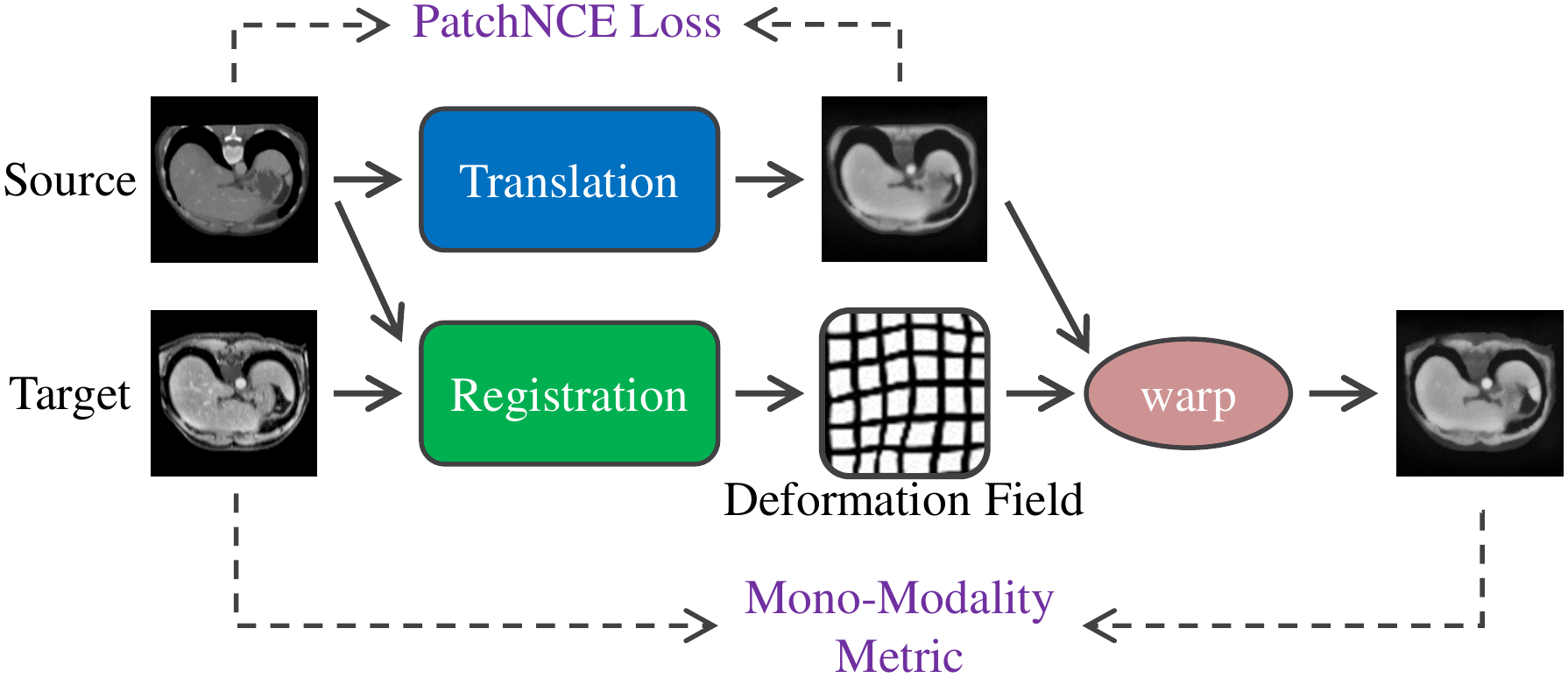}
	\caption{\textbf{Method overview.} Our translation-based registration method learns a cross-modality translation, mapping between the two modalities, which enables the training of the registration network with a mono-modality metric. The source image is warped to align with the target image by the deformation field. We use a PatchNCE loss to encourage the translation network to preserve object shapes and design a pixel loss as the mono-modality metric. The pixel loss not only enables the appearance transfer but also measures the image dissimilarity, which accounts for the training of both translation and registration networks simultaneously.}
	\label{fig1}
\end{figure}

Given the recent success of multi-modal image translation~\cite{huang2018multimodal,park2020contrastive}, an alternative solution for addressing multi-modal registration is to convert the problem to a simpler unimodal task using an image-to-image (I2I) translation framework~\cite{qin2019unsupervised}. Specifically, translation-based methods use Generative Adversarial Network (GAN) mode to translate images from source modality to target modality. And the GAN consists of a generator and a discriminator, where the generator learns to generate plausible data and the discriminator penalizes the generator for producing unrealistic results. With the GAN mode, the registration network can be trained with unimodal similarity metrics. However, this GAN-based image translation tends to produce shape inconsistency and artificial anatomical features which will, in turn, deteriorate the performance of the registration~\cite{arar2020unsupervised,xu2020adversarial}. More specifically, different modalities have very distinct geometric variances caused by the shape of the imaging bed, the imaging protocols of the scanner, and the field of view. We refer to these variances as \textbf{``domain-specific deformations''}~\cite{wang2021dicyc}. We argue that the inconsistency and artifacts are introduced by the discriminator that mistakenly encodes domain-specific deformations as indispensable appearance features and encourages the generator to reproduce the deformations. This tends to create unnecessary difficulty for registration tasks. This paper shows that we can improve the performance of multi-modal image registration by removing the discriminator in image-to-image translation.

In this work, we propose a novel translation-based unsupervised registration approach for aligning multi-modal images. Our main idea is to reduce the inconsistency and artifacts of the translation by removing the discriminator as we discussed above. Specifically, we replace the GAN-based translation network with our discriminator-free translation network (shown in Figure \ref{fig1}). The presented translation network incorporates a patchwise contrastive loss (PatchNCE~\cite{park2020contrastive}) to maintain shape consistency during translation and a pixel loss to integrate the output with the target appearance. Additionally, two novel loss terms are also proposed, one is local alignment loss and the other is global alignment loss. The local alignment loss captures detailed local texture information by modeling the detail-rich image patches. The global alignment loss focuses on the overall shape for the detail-missing generated images. The proposed translation-based registration method coupled with these two losses can achieve local and global alignment and yield more accurate deformation fields.

The main contributions of our work can be summarized as follows:
\begin{itemize}
    \item We present a discriminator-free I2I translation mode to replace the original GAN-based I2I mode and achieve accurate registration.
    \item We design a contrastive PatchNCE loss in our translation-based registration model as a shape-preserving constraint.
    \item We also propose a local and a global alignment losses to further improve the registration performance.
\end{itemize}

\section{Related Work}
In recent years, many translation-based multi-modal registration approaches have been proposed. They commonly follow a registration-by-translation framework: an I2I translation network is first trained to synthesize fake images with the target appearance, and then mono-modality metrics can be used in the target domain. In unsupervised translation, cycle-consistent generative adversarial networks (CycleGAN~\cite{zhu2017unpaired}) are widely adopted. However, cycle consistency leads to multiple solutions, which means that the translated images can not maintain the anatomical structure of source images and may contain artifacts~\cite{kong2021breaking}. With inaccurate image translation, the performance of multi-modal registration tends to be degraded. To address this difficulty, other approaches beyond CycleGAN have been proposed.

Qin et al.~\shortcite{qin2019unsupervised} use image disentanglement to decompose images into common domain-invariant latent shape features and domain-specific appearance features. Then the latent shape features of both modalities are used to train a registration network. But this method still relies on cycle consistency and a GAN mode, which inevitably introduces inconsistency and hampers the process of registration.

Arar et al.~\shortcite{arar2020unsupervised} attempt to force the translation and the registration steps to be commutative, which can implicitly encourage the translation network to be structure-preserving. The structure-preserving translation network allows the use of simple mono-modality metrics for training a registration network. However, their method is still GAN-based, which means the structure consistency will be affected by the presence of the discriminator.

Closest to our work, Casamitjana et al.~\shortcite{casamitjana2021synth} propose a synthesis-by-registration method, which is different from the previous registration-by-synthesis methods. Their approach is made up of two stages: stage one is training a registration network on pairs of images from the target domain with data augmentation, and stage two is training an I2I network while freezing the parameters of the registration network. However, their registration network is first trained on the images from the target modality instead of images from the two modalities, which may guide the registration network to generate an unrealistic deformation field. And both the registration network and translation network are used in test time in their scheme. On the contrary, our method is a joint framework with end-to-end optimization and only the registration network is needed in test time, which leads to a more reliable deformation field.

\section{Method}

\begin{figure*}
	\centering
	\includegraphics[scale=0.44]{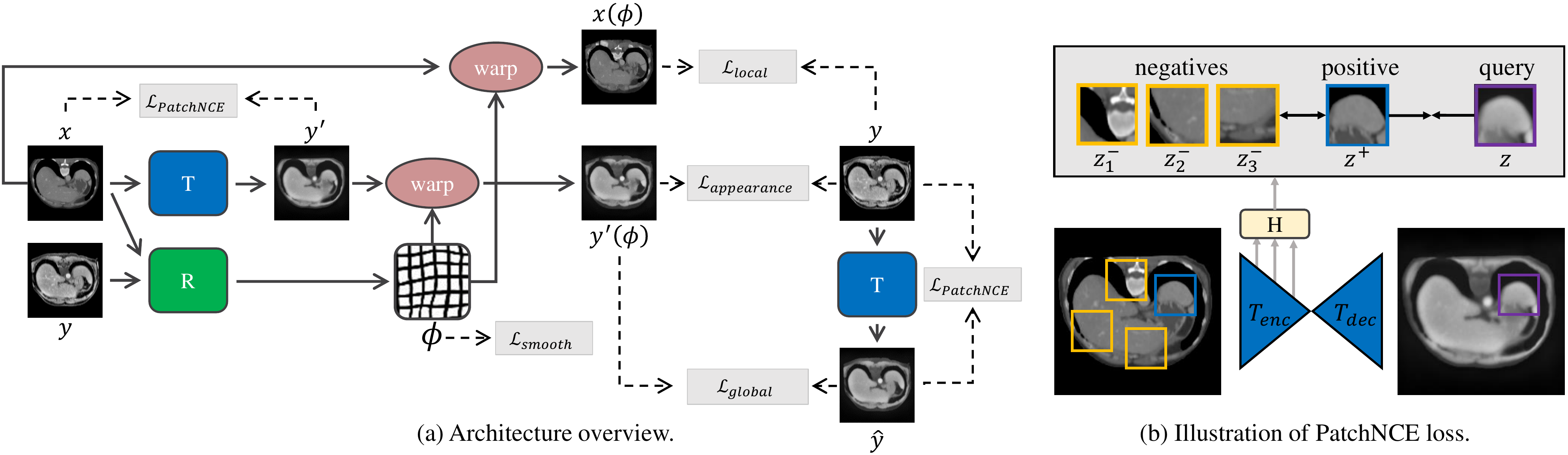}
	\caption{\textbf{An overview of the proposed method.} (a) Architecture overview. Our model consists of two components: a registration network $R$ and a discriminator-free translation network $T$. The two networks are jointly trained in an end-to-end manner. In our context, the pixel loss $\mathcal{L}_\mathrm{appearance}$ is the mono-modality metric computed in the target modality. Based on the architecture presented in Figure \ref{fig1}, we add two novel loss terms $\mathcal{L}_\mathrm{local}$ and $\mathcal{L}_\mathrm{global}$ to achieve local and global alignments between $x(\phi)$ and $y$. (b) Illustration of the PatchNCE loss.}
	\label{fig2}
\end{figure*}
In this work, we propose an end-to-end learning framework for registering multi-modal image pairs in a fully unsupervised manner. Our core idea is to replace GAN-based I2I translation with our novel discriminator-free I2I translation, shown in Figure \ref{fig1}. The PatchNCE loss maintains shape consistency and the pixel loss enables the appearance transfer. With the proposed discriminator-free I2I translation, the multi-modal registration task is converted into a unimodal one.

Our method consists of two components: (1) a deformable registration network, and (2) a discriminator-free translation network for mapping images from source domain to target domain and reconstructing images from the target domain. The two components are trained jointly, and only the registration network is used in test time. The pipeline of our method is depicted in Figure \ref{fig2}(a). 

We denote the registration network as $ R $ and the translation network as $ T $, as in Figure \ref{fig2}. Let  $ \mathcal{X} $ and $ \mathcal{Y} $ denote two paired image domains, where $ \mathcal{X} $ is source domain and $ \mathcal{Y} $ is target domain. Pairing means that each image $ x \in \mathcal{X} $ has a corresponding unaligned image $ y \in \mathcal{Y} $ representing the same anatomical structure. The process of registration is to find a deformation field that aligns the source image $ x $ to the target image $ y $ accurately. Given an image pair $(x,y)$ as input, $ R $ learns to predict a deformation field $ \phi $, which describes how to non-rigidly align $ x $ to $ y $. Meanwhile, $ T $ takes $ x $ and $y$ as the inputs and outputs target-modality images $ y' $ and $ \hat y $, where $y'=T(x)$ and $\hat y=T(y)$. $y'$ is the translated image, which has similar appearance as images in domain $\mathcal{Y}$. $\hat y$ is the reconstructed image of $y$. The PatchNCE loss $\mathcal{L}_\mathrm{PatchNCE}$ is employed to force $y'$ and $\hat y$ to keep the orginial structure of $x$ and $y$, respectively. $x(\phi)$ and $y'(\phi)$ are warped images from the source image $x$ and the translated image $y'$. The pixel loss $\mathcal{L}_\mathrm{appearance}$ enables the appearance transfer of network $T$ between $y'(\phi)$ and $y$. In addition, local alignment loss $\mathcal{L}_\mathrm{local}$ and global alignment loss $\mathcal{L}_\mathrm{global}$ are employed to further improve the registration performance.

\subsection{Registration Network}
The registration network $R$ takes an image pair $(x,y)$ as an input and outputs a deformation field $ \phi=R(x,y) $. The warped image $ x(\phi) $ is aligned with $ y $. In a two-dimensional setting, the deformation field is a matrix of 2D vectors, indicating the moving direction for every pixel in the source image $ x $. To generate smooth deformation fields and penalize the tendency of overly distorting the deformed image $x(\phi)$, we adopt an $L_{2}$-norm of the gradients of the deformation field as the regularization term~\cite{hoopes2021hypermorph}, which is denoted as $\mathcal{L}_{smooth}$. Formally, the loss at pixel $v=(i,j)$ is given by:
\begin{equation}
    \mathcal{L}_{smooth}(\phi,v)=\sum_{u \in N(v)} \left\|\phi(u)-\phi(v)\right\|_2 ,
\end{equation}
where $N(v)$ denotes a set of neighbor pixels of the $v$.

\subsection{Discriminator-Free Translation Network}
Our translation network $T$ takes images from the source domain $\mathcal{X}$ and outputs translated images that have similar appearance as images in the target domain $\mathcal{Y}$. We divide our translation network $T$ into two components, an encoder $T_{enc}$ followed by a decoder $T_{dec}$, shown in Figure \ref{fig2}(b). $T_{enc}$ extracts shape-related features, while $T_{dec}$ learns to perform shape-preserving modality translation with those features. Given the input $x$, $T_{enc}$ and $T_{dec}$ jointly generate the output $y'=T(x)=T_{dec}\left(T_{enc}(x)\right)$.

A key task of our method is to train the I2I translation network $T$ to translate images without the discriminator. If the output from $T$ is shape-preserving, which implies $T$ cannot deform the anatomical structure, the alignment task will be done solely by registration network $R$. Therefore, we propose the contrastive PatchNCE loss to enforce the shape consistency and the pixel loss to enable the appearance transfer from the source modality to the target modality.

\textbf{PatchNCE Loss.}
The PatchNCE loss maximizes the mutual information between input image patches and output image patches, which is based on a noise contrastive estimation framework~\cite{oord2018representation}. It makes every output patch similar to the corresponding input patch, while different from the other patches within the input. We use a ``query'' to refer to an output patch and ``positive'' for the corresponding input patch and ``negatives'' for the noncorresponding input patches, shown in Figure \ref{fig2}(b). For similarity measurement, the encoder part $T_{enc}$ of our translation network with additional two-layer multi-layer perceptron (MLP) is used to map image patches to embedded vectors. Specifically, the query, positive, and $N$ negatives are embedded to $K$-dimensional vectors $z$, $z^{+}\in \mathbb{R}^{K}$ and $z^{-}\in \mathbb{R}^{N \times K}$, respectively. $z_{n}^{-}\in \mathbb{R}^{K}$ denotes the n-th negative patch. We convert the similarity measurement to an (N+1)-way classification problem, where the similarity between the query and other samples are expressed as logits. The cross-entropy loss for multi-class classification is calculated, representing the probability of the positive being selected over $N$ negatives. The formula is given by:
\begin{align}
    & \ell\left( {z}, {z}^{+}, {z}^{-} \right)=\nonumber\\
    & -\log \left[\frac{\exp \left({z} \cdot {z}^{+} / \tau\right)}{\exp \left( {z} \cdot {z}^{+} / \tau\right)+\sum_{n=1}^{N} \exp \left({z} \cdot {z}_{n}^{-} / \tau\right)}\right],
\end{align}
where $\tau$ is the temperature parameter set to 0.07 in our experiments.

Given an input image, the encoder part $T_{enc}$ will generate multilayer hidden features, which form a feature stack. A spatial location in a layer of the feature stack represents a patch of the input image, with deeper layers corresponding to bigger patches. Let $L$ denote the number of extracted layers from the feature stack. A MLP $H_{l}$ with two layers is used to map the selected encoder features to embedded representations $\left \{z_{l} \right \}_{L}=\left\{H_{l}\left(T_{enc }^{l}(x)\right)\right\}_{L}$, where $T_{enc}^{l}$ represents the features of the $l$-th selected layer and $l \in \left\{ 1,2,\ldots,L \right\}$. Let $s\in \left\{ 1,\ldots,S_{l} \right\}$, where $S_{l}$ represents the number of spatial locations in each $T_{enc}^{l}$. For each spatial location $s$, the corresponding embedded code is referred to as ${z}_{l}^{s} \in \mathbb{R}^{K}$, and the features at any other locations are denoted by $ {z}_{l}^{S \backslash s} \in \mathbb{R}^{\left(S_{l}-1\right) \times K} $. Similarly, the corresponding embedded representations of output image $y'$ are $\left \{z'_{l} \right \}_{L}=\left\{ H_{l}\left(T_{enc }^{l}\left(G\left(x\right) \right)\right) \right\}_{L}$.

With features from multiple layers of the encoder, patchwise noise contrastive estimation can be applied on multiple scales. Multilayer PatchNCE loss is given by:
\begin{equation}
\mathcal{L}_{\mathrm{PatchNCE}}(T,H,X)=\mathbb{E}_{x} \sum_{l=1}^{L} \sum_{s=1}^{S_{l}} \ell\left({z'}_{l}^{s}, {z}_{l}^{s}, {z}_{l}^{S \backslash s}\right).
\end{equation}

In addition, $\mathcal{L}_{\mathrm{PatchNCE}}$ is also used on images from target domain $\mathcal{Y}$, which acts as reconstruction loss. Through the loss $\mathcal{L}_{\mathrm{PatchNCE}}(T,H,Y)$, network $T$ outputs the reconstruction $\hat y=T(y)$.

\textbf{Pixel Loss.}
Contrastive loss is effective to preserve the shape of the input image $x$. However, without adversarial loss, we still need to maximize the appearance similarity between a translated image $y'$ and the target domain $\mathcal{ Y}$ with a pixel loss. We define the pixel loss with $L_{1}$-norm as:
\begin{equation}
    \mathcal{L}_\mathrm{{appearance}}(T,R)=\left\|y'(\phi)-y\right\|_1 ,
\end{equation}
where  $y'(\phi)$ indicates the warped image of $y'$.

$\mathcal{L}_\mathrm{appearance}$ explicitly penalize the absolute intensity differences between $y'(\phi)$ and $y'$. The combination of $\mathcal{L}_\mathrm{appearance}$ and $\mathcal{L}_\mathrm{PatchNCE}$ leads to a discriminaotr-free and shape-preserving translation. Note that by minimizing the two losses, the registration network $R$ is trained jointly to predict a deformation field $\phi$, which aligns $y'$ to $y$.

\subsection{Local and Global Alignment}
To enable $R$ to learn the alignment at the local (patch) level, we propose a variant of the PatchNCE loss. Similarly, we refer to a patch of the warped source image $x(\phi)$ as a ``query'', while ``positive'' and ``negatives'' are corresponding and noncorresponding patch(es) within the target image $y$, respectively. The warped source image and the target image are mapped to embedded vectors $\left \{q_{l} \right \}_{L}=\left\{H_{l}\left(T_{enc }^{l}(x(\phi))\right)\right\}_{L}$ and $\left \{p_{l} \right \}_{L}=\left\{H_{l}\left(T_{enc }^{l}(y)\right)\right\}_{L}$, respectively. The patchwise noise contrastive estimation is computed between the embedded vectors $\left\{q_{l} \right \}_{L}$ and $\left\{p_{l} \right \}_{L}$, which is different from the original PatchNCE. For clarity, we denote this variant loss as $\mathcal{L}_{\mathrm{local}}$:
\begin{equation}
\mathcal{L}_{\mathrm{local}}(R)=\mathbb{E}_{x,y} \sum_{l=1}^{L} \sum_{s=1}^{S_{l}} \ell\left({q}_{l}^{s}, {p}_{l}^{s}, {p}_{l}^{S \backslash s}\right).
\end{equation}

\begin{figure}
	\centering
	\includegraphics[scale=0.35]{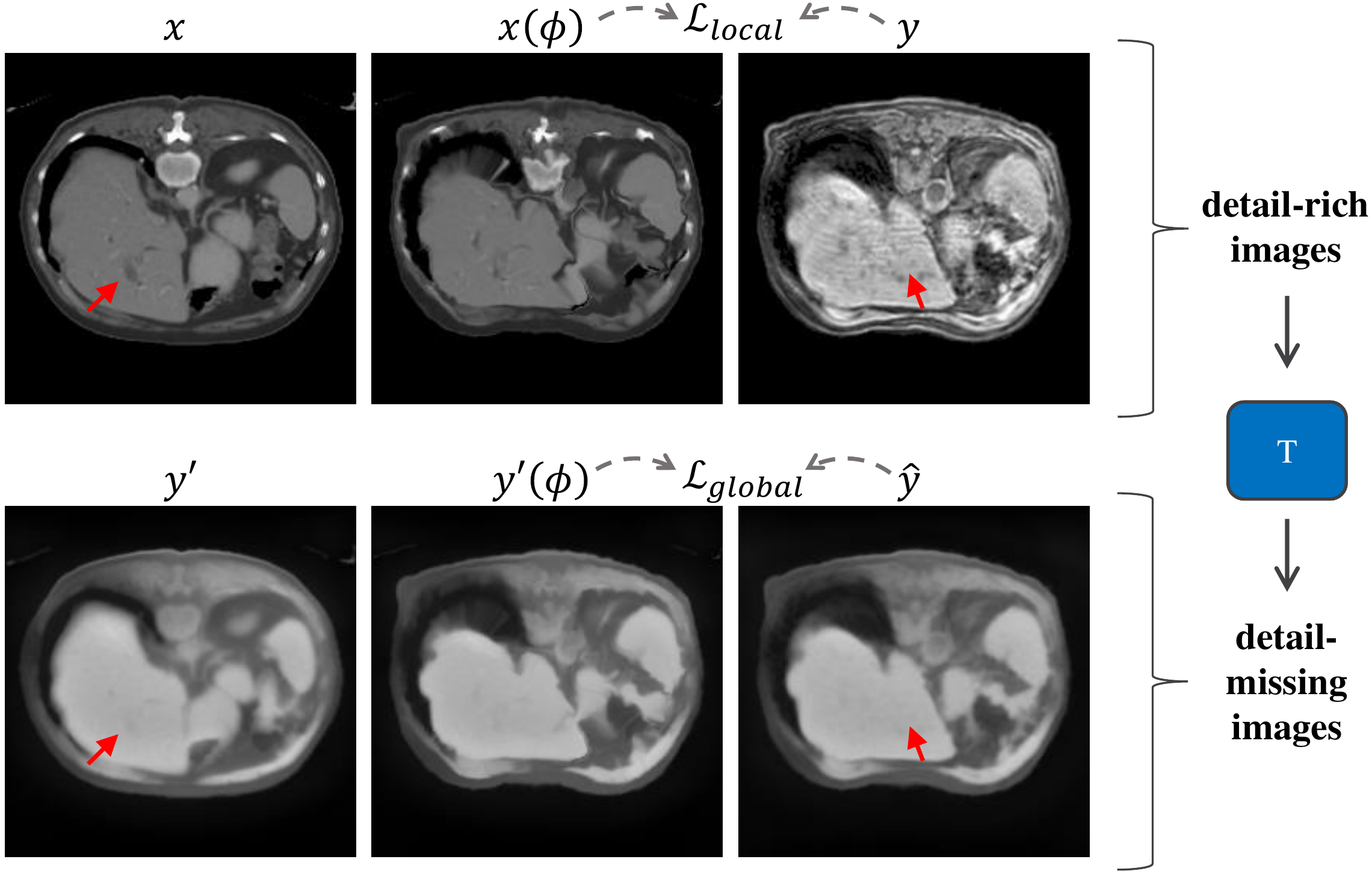}
	\caption{\textbf{Illustration of local and global alignment losses.} Local information such as texture is provided by the detail-rich original image patches (top row). The generated image pair $(y',\hat y)$ only retains the basic shapes and ignores image texture and details, leading to the focus on global information such as overall shape (bottom row).}
	\label{fig4}
\end{figure}

Applied on the cross-modality image patches, $\mathcal{L}_{\mathrm{local}}$ encourages the registration network $R$ to learn local alignment as shown in Figure \ref{fig4}. Note that the images produced by our discriminator-free translation network do not contain image texture information. This is beneficial to the extraction of global information, shown in Figure \ref{fig4}. Inspired by this, we further propose a global alignment loss: 
\begin{equation}
    \mathcal{L}_\mathrm{{global}}(T,R)=\left\|y'(\phi)-{\hat{y}}\right\|_1.
\end{equation}

Minimizing $\mathcal{L}_\mathrm{global}$ leads to similar style between the generated $y'$ and $\hat y$. Meanwhile, the registration network $R$ learns a deformation field $\phi$ to best align $y'$ to $\hat y$.

\subsection{Final Objective}
Our final objective is as follows:
\begin{align}
    \mathcal{L}= \lambda_{P} \cdot \mathcal{L}_{\mathrm{PatchNCE}}(T,H,X) 
    + \lambda_{P} \cdot \mathcal{L}_{\mathrm{PatchNCE}}(T,H,Y) \nonumber\\
    + \lambda_{A} \cdot \mathcal{L}_\mathrm{{appearance}}(T,R) 
    + \lambda_{L} \cdot \mathcal{L}_{\mathrm{local}}(R) \nonumber\\
    + \lambda_{G} \cdot \mathcal{L}_\mathrm{{global}}(T,R),
\end{align}
where we set $\lambda_{P}=0.25$, $\lambda_{A}=1$, $\lambda_{L}=0.25$ and $\lambda_{G}=1$ in our experiments.

\section{Experiments}
\subsection{Datasets}
We evaluated our proposed method on two public datasets. Both of them are obtained from MICCAI Learn2Reg 2021 challenge~\cite{hering2021learn2reg}, which is a comprehensive registration challenge covering different anatomical structures and modalities. Specifically, the first dataset is for the task of CT-MR thorax-abdomen intra-patient registration, and the other one is for the task of CT lung inspiration-expiration registration.

\textbf{Thorax-Abdomen CT-MR Dataset.} This dataset contains 16 pairs of CT and MR abdomen scans. The annotations on all scans are manual and automatic segmentations of multiple organs. Each scan is a 3D volume in size of $192 \times 160 \times 192$ with 2 $mm$ voxel spacing. After preprocessing the data, e.g., coarse affine registration using the Elastix toolbox~\cite{marstal2016simpleelastix}, we randomly split the dataset into 10/2/4 pairs for train/validation/test, and central 90 slices containing organs in each scan are extracted for our experiments. All slices are padded and resized into $256 \times 256$.

\textbf{Lung CT Inspiration-Expiration (Insp-Exp) Dataset.} The dataset consists of 30 pairs of CT lung scans. All lungs are segmented automatically or manually. The challenge in this dataset is to estimate the underlying breathing motion between inspiration and expiration. Each scan is a 3D volume in size of $192 \times 192 \times 208$ with resolution of 1.75 $mm$. In our experiments, we randomly divide the 30 pairs into 20/4/6 for train/validation/test, and on each volume, we extract middle 100 slices. And each slice is resized into $256 \times 256$. To simulate a multi-modal registration task with this single-modal dataset, we synthesize a new modality using the intensity transformation function $\cos{\left(I \cdot \pi \middle/ 255 \right)}$, Gaussian blurring with the kernel size of $3\times 3$ and random elastic deformation sequentially as proposed in~\cite{qin2019unsupervised}. Note that the intensity transformation function is only applied to the foreground of each slice. Deformation fields are predicted between real CT inspiration slices and their corresponding synthesized expiration slices.

\subsection{Implementation Details}
We implement our model based on the framework and implementation of CUT~\cite{park2020contrastive}. The translation network $T$ is a Resnet-based architecture with 9 residual blocks. Our encoder is defined as the first half of the translation network, and five layers of features in the encoder are extracted. The registration network adopts a U-net based architecture with skip connections from contracting path to expanding path~\cite{ronneberger2015u}. For the initialization of networks, we use the Xavier initialization method. Our networks are implemented in PyTorch and all the experiments were conducted on GeForce RTX 2080 Ti. We use Adam optimizer to train our model for 300 epochs with parameters $lr=0.0002 $, $\beta_{1}=0.5$ and $\beta_{2}=0.999$. Linear learning rate decay is activated after 200 epochs.

\subsection{Evaluation}
\textbf{Metrics.} For the thorax-abdomen dataset, we directly use multi-organ segmentation masks to evaluate the registration accuracy. Dice similarity coefficient (DSC~\cite{dice1945measures}) and $95\%$ percentile of Hausdorff distance (HD95~\cite{huttenlocher1993comparing}) are computed between multi-organ masks of a warped source image and its target image. Similarly, DSC and HD95 are calculated between the provided lung masks for the evaluation of the lung CT dataset. DSC is used to measure the accuracy of registration, while HD95 measures reliability. A higher DSC and lower HD95 indicate a better performance of the registration model.

\textbf{Baselines.} We compare our method against several recent state-of-the-art multi-modal registration methods and some other well-established methods. Specifically, the competing methods are: (1) \textbf{Affine}, affine registration based on the normalized mutual information using the Elastix toolbox; (2) \textbf{MIND}, a VoxelMorph architecture~\cite{balakrishnan2019voxelmorph} with similarity metric MIND; (3) \textbf{CGAN}, a CycleGAN, which is pre-trained on unpaired images, combines with the VoxelMorph registration network using mono-modal similarity metric NCC; (4) \textbf{RGPT}, a multi-modal registration model via geometry preserving translation~\cite{arar2020unsupervised}; (5) \textbf{SbR}, a recent synthesis-by-registration model based on contrastive learning~\cite{casamitjana2021synth}. For ablation study, we propose three variants of our model by removing the loss terms one by one: ours w/o $\mathcal{L}_\mathrm{local}$, ours w/o $\mathcal{L}_\mathrm{global}$, and ours w/o $\mathcal{L}_\mathrm{local}$ and  $\mathcal{L}_\mathrm{global}$ (ours w/o $l$ \& $g$). To investigate if the inaccuracy is introduced by the discriminator, we also study a variant with a discriminator plugged into the translation network in our model (ours w/ $dis$).

\begin{table}
	\centering
	\scalebox{0.73}{
	\begin{tabular}{lrrrr}
		\toprule
		&\multicolumn{2}{c}{CT $\rightarrow$ MR} &\multicolumn{2}{c}{MR $\rightarrow$ CT}\\
		\cmidrule{2-5}
		
		Method & DSC $\uparrow$ & HD95 $\downarrow$ & DSC $\uparrow$ & HD95 $\downarrow$\\
		\midrule
		Affine & $0.649(0.031)$ & $14.141(2.126)$ & $0.655(0.031)$ & $14.009(2.082)$\\
		MIND & $0.666(0.039)$ & $14.102(2.446)$ & $0.682(0.024)$ & $14.124(2.320)$\\
		CGAN & $0.698(0.027)$ & $13.798(2.097)$ & $0.695(0.018)$ & $13.660(1.782)$\\
		RGPT & $0.713(0.022)$ & $15.844(2.284)$ & $0.717(0.008)$ & $14.706(1.635)$\\
		SbR & $0.735(0.016)$ & $12.761(1.851)$ & $0.746(0.026)$ & $12.672(2.090)$\\
		\midrule
		Ours & $\textbf{0.772}(0.025)$ & $\textbf{12.137}(2.392)$ & $\textbf{0.784}(0.025)$ & $\textbf{11.889}(2.207)$\\
		w/o $\mathcal{L}_\mathrm{local}$ & $0.757(0.027)$ & $12.361(2.609)$ & $0.763(0.022)$ & $12.405(2.045)$\\
		w/o $\mathcal{L}_\mathrm{global}$ & $0.763(0.026)$ & $12.453(2.372)$ & $0.770(0.023)$ & $12.232(2.045)$\\
		w/o $l$ \& $g$ & $0.753(0.021)$ & $12.742(2.384)$ & $0.762(0.022)$ & $12.378(1.952)$\\
		w/ $dis$ & $0.764(0.025)$ & $12.470(2.311)$ & $0.769(0.024)$ & $12.132(2.076)$\\
		\bottomrule
	\end{tabular}}
	\caption{Evaluation of bi-directional multi-modal registration on the Thorax-Abdomen CT-MR dataset in terms of DSC, HD95 (and standard deviation in parentheses).}
	\label{table1}
\end{table}

\begin{table}
	\centering
	\scalebox{0.73}{
		\begin{tabular}{lrrrr}
			\toprule
			&\multicolumn{2}{c}{insp $\rightarrow$ exp} &\multicolumn{2}{c}{exp $\rightarrow$ insp}\\
			\cmidrule{2-5}
			
			Method & DSC $\uparrow$ & HD95 $\downarrow$ & DSC $\uparrow$ & HD95 $\downarrow$\\
			\midrule
			Affine & $0.850(0.050)$ & $13.985(2.570)$ & $0.848(0.053)$ & $13.746(2.374)$\\
			MIND & $0.899(0.057)$ & $11.476(1.656)$ & $0.934(0.047)$ & $9.643(2.973)$\\
			CGAN & $0.879(0.053)$ & $12.989(1.730)$ & $0.913(0.039)$ & $12.038(2.298)$\\
			RGPT & $0.914(0.034)$ & $12.578(1.158)$ & $0.947(0.030)$ & $11.573(1.351)$\\
			SbR & $0.924(0.049)$ & $8.866(2.513)$ & $0.946(0.037)$ & $6.739(2.780)$\\
			\midrule
			Ours & $\textbf{0.938}(0.025)$ & $\textbf{7.737}(1.791)$ & $\textbf{0.966}(0.019)$ & $\textbf{4.367}(1.048)$\\
			w/o $\mathcal{L}_\mathrm{local}$ & $0.930(0.024)$ & $8.874(1.773)$ & $0.963(0.015)$ & $4.402(1.410)$\\
			w/o $\mathcal{L}_\mathrm{global}$ & $0.932(0.029)$ & $8.789(1.936)$ & $0.963(0.019)$ & $5.382(2.409)$\\
			w/o $l$ \& $g$ & $0.910(0.031)$ & $11.714(1.843)$ & $0.957(0.022)$ & $5.642(2.973)$\\
			w/ $dis$ & $0.929(0.032)$ & $8.372(2.118)$ & $0.959(0.017)$ & $6.232(1.601)$\\
			\bottomrule
	\end{tabular}}
	\caption{Evaluation of bi-directional multi-modal registration on the Lung Insp-Exp dataset in terms of DSC, HD95 (and standard deviation in parentheses).}
	\label{table2}
\end{table}

\begin{figure*}
	\centering
	\includegraphics[scale=0.35]{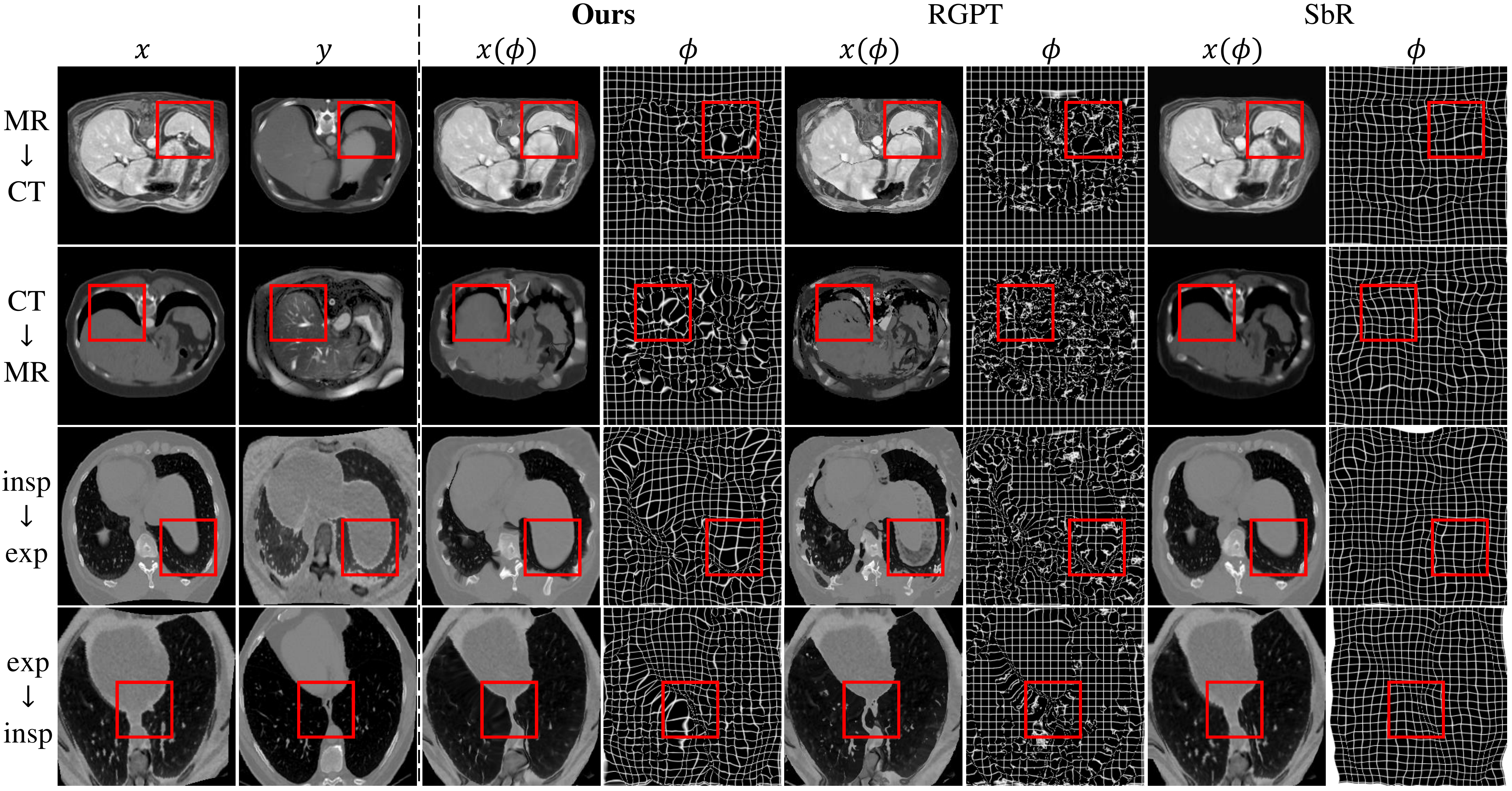}
	\caption{\textbf{Visualization results of our method against other methods.} The unaligned image pair $(x,y)$ is shown in column 1-2 (on the left of the dotted line). $x$ denotes the source image and $y$ denotes the target image. We show the registration results of three methods: Ours, RGPT and SbR. RGPT and SbR are the most recent state-of-the-art methods. Each registration method occupies two columns, the first column showing the warped source image $x(\phi)$ and the second showing the deformation field $\phi$.}
	\label{fig3}
\end{figure*}
\subsection{Results}
The quantitative results on the two datasets are summarized in Table \ref{table1} and Table \ref{table2}. We compare the proposed method with the other five methods by measuring the registration accuracy with the DSC and HD95. Our method consistently outperforms other competing methods. Our registration network can predict more accurate deformation fields, even when there exists significant shape deformation and style difference between source images and target images. In addition, Figure \ref{fig3} displays four examples of the warped source images and their corresponding deformation fields. The deformation fields generated by the RGPT method are corrupted by noise and jitter (the red boxes in column 5-6 in Figure \ref{fig3}). This explains its inferior performance in HD95. Even though RGPT can achieve a good boundary alignment, the image quality is degraded after the registration. On the contrary, SbR produces relatively smooth deformation fields but fails to deal with large-scale deformations (the red boxes in column 7-8 in Figure \ref{fig3}). This is because the registration network of SbR is pre-trained on the images from the target modality with randomly generated deformation fields and its parameters are frozen once the pre-training is done. Different from the above methods, ours achieves the most accurate boundary alignment by capturing both local and global information without any discriminator. (the red boxes in column 3-4 in Figure \ref{fig3}).

\subsection{Ablation Study}
The primary objective of our work is to achieve accurate multi-modal registration. Therefore, we design three variants of the proposed model to investigate the impact of $\mathcal{L}_\mathrm{local}$ and $\mathcal{L}_\mathrm{global}$. The results in Table $1$ and $2$ show that, $\mathcal{L}_\mathrm{local}$ and $\mathcal{L}_\mathrm{global}$ are complementary, since the proposed method outperform the three ablated versions. What's more, with only one of the two losses, the registration performance still can be improved. To investigate whether the discriminator will introduce inconsistency and degrade the performance of the registration, we design a variant, ours w/ $dis$. As can be seen in the last row of Table $1$ and $2$, the registration performance becomes worse than the proposed method.

\subsection{Conclusion}
We propose a novel discriminator-free and shape-preserving translation network for multi-modal registration, taking advantage of contrastive learning. The registration network successfully gets rid of inconsistency and artifacts introduced by the discriminator. The contrastive loss ensures shape consistency while the pixel loss enables the appearance transfer. Furthermore, we leverage a local and a global alignment losses to achieve local and global alignment, improving the registration accuracy. Finally, we evaluate the proposed method on two open datasets, and show that it outperforms the state-of-the-art methods.

\section*{Acknowledgments}
This work is supported in part by the Natural Science Foundation of Guangdong Province (2020A1515010717), NSF-1850492 and NSF-2045804.
\small
\bibliographystyle{named}
\bibliography{ijcai}

\end{document}